\documentclass[letterpaper, 10 pt, conference]{ieeeconf}  

\IEEEoverridecommandlockouts                              

\usepackage{siunitx}

\usepackage[whole]{bxcjkjatype} 
\usepackage[space]{cite}

\newcommand\myswitchlang[2]{
  \ifx\mypaperlang\empty
  #1
  \else
  #2
  \fi
}
\def\mypaperlang{en}

\usepackage{graphicx}
\graphicspath{{figs/}}
\usepackage{amsmath}
\usepackage{amssymb}
\usepackage{algorithm}
\usepackage{algpseudocode}
\usepackage{times}

\usepackage{adjustbox}
\usepackage{multirow}
\newcommand{\figref}[1]{Fig.~\ref{figure:#1}}
\newcommand{\tabref}[1]{Table~\ref{table:#1}}

\title{\LARGE \bf
Integrative Wrapping System for a Dual-Arm Humanoid Robot
}

\author{Yukina Iwata$^{1}$, Shun Hasegawa$^{1}$, Kento Kawaharazuka$^{1}$, Kei Okada$^{1}$, and Masayuki Inaba$^{1}$
\thanks{$^{1}$The authors are with the Department of Mechano-Informatics, Graduate School of Information Science and Technology, The University of Tokyo, 7-3-1 Hongo, Bunkyo-ku, Tokyo, 113-8656, Japan. [y-iwata, hasegawa, kawaharazuka, k-okada, inaba]@jsk.imi.i.u-tokyo.ac.jp}}

\begin{document}

\maketitle
\thispagestyle{empty}
\pagestyle{empty}

\begin{abstract}

\myswitchlang
    {
    紙や布を扱う柔軟物操作はロボットマニピュレーションにおいて大きな研究課題である。従来は特定の動作を可能にするハードウェア開発や、sim-to-realや学習を用いた紙を折る一動作の実現などが行われてきたが、ヒューマノイドロボットでの実現や連続的でマルチステップな動作を可能にするシステムの提案にはあまりいたっていない。
    紙で物体を覆いテープを貼る包装動作は、折り紙や布の折りたたみに比べて扱う物体が増え、また立体性が増すことにより、従来のマニピュレーション研究に比べて動作の多様性、煩雑性が増す。本研究では包装で扱う各物体の特徴をもとに必要な情報を整理し記号化する。また、ヒューマノイドによる包装の一連動作を可能にするハードウェア構成とマニピュレーション方法、認識システムを一般化する。システムには紙のテンションに着目したアドミッタンス制御を伴う操作や点群を用いた状態評価など、三次元的に柔軟物体を扱うためのシステムが含まれる。最後に形状の異なる物体の包装を実験し、提案システムの一般性や有効性を示す。}
    {
    Flexible object manipulation of paper and cloth is a major research challenge in robot manipulation. Although there have been efforts to develop hardware that enables specific actions and to realize a single action of paper folding using sim-to-real and learning, there have been few proposals for humanoid robots and systems that enable continuous, multi-step actions of flexible materials.
    Wrapping an object with paper and tape is more complex and diverse than traditional manipulation research due to the increased number of objects that need to be handled, as well as the three-dimensionality of the operation. In this research, necessary information is organized and coded based on the characteristics of each object handled in wrapping. We also generalize the hardware configuration, manipulation method, and recognition system that enable humanoid wrapping operations. The system will include manipulation with admittance control focusing on paper tension and state evaluation using point clouds to handle three-dimensional flexible objects. Finally, wrapping objects with different shapes is experimented with to show the generality and effectiveness of the proposed system.}

\end{abstract}

\section{INTRODUCTION}
\begin{figure}[h]
  \centering
  \includegraphics[width=0.9\columnwidth]{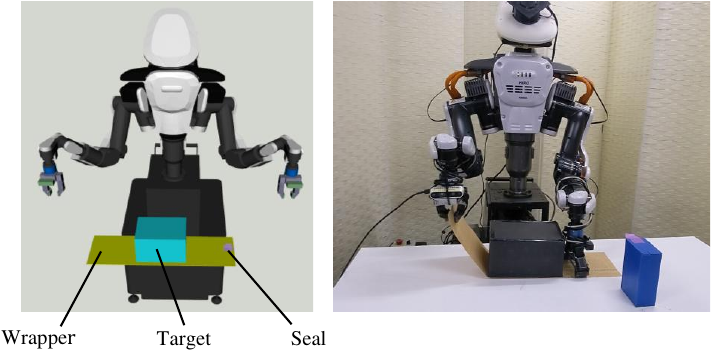}
  \caption{Wrapping with a dual-armed humanoid robot. The wrapping operation deals with three things: the object to be wrapped, the object that wraps the object, and the object that seals the object. In this paper, these three are represented as Target, Wrapper, and Seal, respectively.}
  \label{figure:wrapping}
\end{figure}

\myswitchlang
    {紙や布、紐は日常生活や産業において欠かせない素材であり、ロボットによる生活支援や自動化の実現にはこれらの柔軟物体をハンドリングできるロボットの開発は不可欠である。

      一方、柔軟物体のロボットによるマニピュレーションは難しく、今まで様々な観点から研究が行われている。アプローチの一つとして折り紙や布の折りたたみにおけるある特定の操作を可能にするハードウェアの開発\cite{balkcom2004introducing, balkcom2008robotic, 4399358}がある。また、別のアプローチとして学習\cite{tamei2011reinforcement, tsurumine2019deep, joshi2019framework}による動作の生成の自動化や動作の改善がある。これらの研究の問題点として、汎用的なハードウェアシステムへの応用性や動作の一般化、柔軟物体の連続的動作への応用が挙げられる。これを受け、本研究では汎用性の高い双腕ヒューマノイドロボットを用いた柔軟物体操作を行うためのシステムについて取り組む。柔軟物体操作の中でも扱う物体数が多く、紙や布をめくる動作やテープで紙を固定する動作など複数の動作要素から成り立ち、且つそれらを手順に沿って連続的に行わなければならない包装動作を対象とする。紙をめくり、物体を覆い、紙の固定のためにテープを貼るという包装の一連の動作\figref{wrapping}の達成を本研究の目的とし、それを実現するための対象物体の記号化、ハードウェア設計の整理、認識やマニピュレーション動作の一般化を行い、統合的なシステムを構築する。}
    {Paper, cloth, and string are indispensable materials in daily life and industry, and the development of robots that can handle these flexible objects is essential for the realization of assistance in our daily life and automation by robots.

      On the other hand, the manipulation of flexible objects by robots is complex and has been studied from various perspectives. One approach is the development of hardware that enables specific manipulations in origami and cloth folding \cite{balkcom2004introducing, balkcom2008robotic, 4399358}. Another approach is automating motion generation and improving motion by learning \cite{tamei2011reinforcement, tsurumine2019deep, joshi2019framework}. The problems of these studies include applicability to general-purpose hardware systems, generalization of the behavior, and application to the continuous motion of flexible objects. In response to these issues, this study addresses a system for flexible object manipulation using a versatile dual-armed humanoid robot. Among the flexible object manipulations, the target is a wrapping operation that requires more objects to be manipulated, consists of multiple motion elements such as turning paper or cloth and fixing paper with tape, and must be performed continuously according to a procedure. This research aims to achieve a sequence of wrapping actions such as turning the paper, covering the object, and applying tape to fix the paper. To achieve this, we construct an integrative system that includes symbolization of the target object, organizing hardware requirements, and generalization of recognition and manipulation actions.}

\section{PROBLEM STATEMENT}
\subsection{Advantages of Using Humanoid Robots}
    \myswitchlang
        {ロボットによる柔軟物操作の代表として、紙を折る動作がある。先駆的研究としてCMUのプレートとスロット付きのハードウェア\cite{balkcom2004introducing, balkcom2008robotic}やロボットハンドによる折り紙の研究\cite{4399358}がある。これらは特殊なハードウェアを開発し特定の折り方を可能にしているが、汎用的な折り紙の操作までは至っていない。また、折り紙以外の動作を行うことは難しい。これらを受けて本研究では異なるタスクが動作可能なヒューマノイドロボットを用いる。各手順に必要な認識やマニピュレーションを関数化することで動作を一般化し、複数の動作や物体に対応可能なシステムを構築する。}
        {A representative example of robotic manipulation of flexible objects is the act of folding paper. Pioneering studies include CMU's plate and slotted hardware \cite{balkcom2004introducing, balkcom2008robotic} and origami with a robotic hand \cite{4399358}. These have developed specialized hardware to enable specific folding techniques but have yet to reach the point of general-purpose origami manipulation. In addition, it is difficult to perform operations other than origami. In response to these issues, this study uses a humanoid robot that can perform different tasks. We generalize the behavior by functionalizing the recognition and manipulation required for each procedure and constructing a system that can handle multiple movements and objects.}

\subsection{Flexible Object Manipulation with Three-dimensionality}
\myswitchlang
    {包装の特徴として立体性を伴う柔軟物操作であるという側面が挙げられる。これまで紙をメッシュの集合体やバネダンパモデルと見做してシミュレーションを行うことで変形を記述し、紙を折る動作を行う研究\cite{7354175, 7844086}が行われてきた。しかし、柔軟物を物で覆い、テープなどで固定する包装動作は対象物が増えることにより物体との衝突が増えシミュレーションモデルの構築は困難になる。
      また、折り紙が折り目に沿って集中的に折るのとは異なり、包装動作は対象物の形に合わせるために、素材を大きく湾曲させたり折り目をつけたりするといったより三次元的な紙の配置が求められ、紙が膨らんだりシワになったりしやすい。
      加えて、紙で覆うことにより物体が隠れる部分が現れることにみえるように、物の状態が動作中に変化する。以上のような立体性が増えることによる課題を解決する認識や動作、対象物体の表現方法を盛り込んだシステムを議論する必要がある。}
    {One of the characteristics of wrapping is that it is a flexible material manipulation that involves three-dimensionality. Previous studies have used mesh or spring-damper models to simulate paper deformation and folding \cite{7354175, 7844086}. However, in the case of a wrapping motion in which a flexible object is covered with an object and fixed with tape, the number of collisions with objects increases as the number of objects increases, making it challenging to construct a simulation model.
      Also, unlike origami's focused folding along crease lines, wrapping requires a more three-dimensional arrangement of the paper, such as large curves and creases in the material to accommodate the object's form, which can cause the paper to bulge and wrinkle.
      Moreover, the object's state changes during the operation, as seen in the appearance of parts of the object hidden by the paper covering. It is necessary to discuss a system that incorporates methods of recognition, operation, and representation of the target object to resolve the issues caused by the increased three-dimensionality described above.}

\subsection{Multi-step Procedure of Flexible Objects}
\myswitchlang
    {動作の連続性も包装動作を行ううえで重要である。これまでにもロボットによる柔軟物体を含む物体の把持について、爪を備えた二本指の平行ソフトグリッパによる紙をめくる動作 \cite{yoshimi2012picking}、吸引 \cite{ponraj2019pinch}や電気粘着力 \cite{shintake2016versatile}を用いた柔らかい物体の把持などが行われてきた。また、模倣学習や強化学習を用いたロボットによる布の物体操作 \cite{tamei2011reinforcement,tsurumine2019deep,joshi2019framework}、包装タスクの自動化 \cite{robomech}が行われてきた。これらはある特定の一動作の自動化や改善に注目している。包装は紙や布の把持からテープによる固定などの状態の保持まで各動作要素を一連で行うものである。
      ラッピングは各サブステップが前のステップに依存する逐次的な性質をもった、multi-stepな動作である。
      そのため、両手を使ったマニピュレーションや,連続して動作を行うことを考慮したハードウェアシステムや認識を考える。}
    {The continuity of motion is also essential for the wrapping operation. In the past, grasping of objects, including flexible objects, by robots has been performed, such as paper flipping motion using a two-finger parallel soft gripper with nails \cite{yoshimi2012picking}, suction \cite{ponraj2019pinch} and electro-adhesive \cite{shintake2016versatile} have been used to grasp soft objects. In addition, manipulation of cloth objects by robots using imitation learning and reinforcement learning \cite{tamei2011reinforcement,tsurumine2019deep,joshi2019framework} and automation of wrapping tasks \cite{robomech} have been done. These focus on automating or improving one particular operation. Wrapping is a series of movement elements, from grasping paper or cloth to holding it in place with tape.
      Wrapping is a multi-step operation with sequential nature in which each substep depends on the previous one.
      Therefore, we propose a system that includes hardware and recognition that considers continuous motion and manipulation using both hands.
}

\section{fundamentals of wrapping operation by a robot}

\begin{figure*}[pthb]
  \centering
  \includegraphics[width=0.9\textwidth]{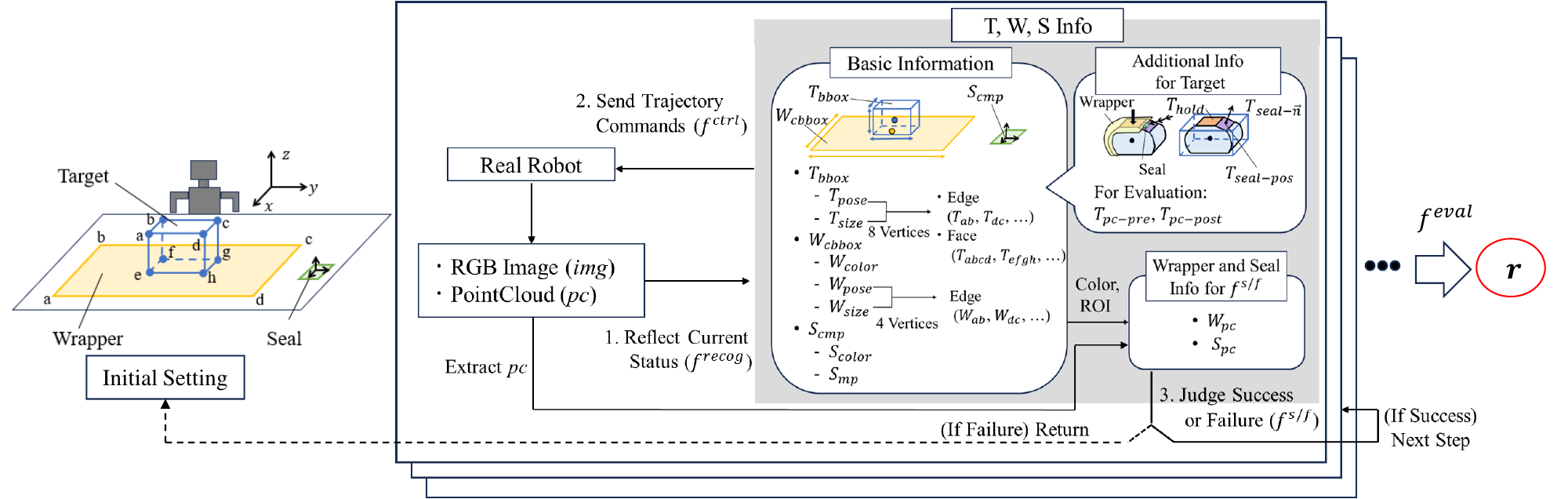}
  \caption{
    Representation of Target, Wrapper, Seal, and the overall system. The basic information of Target, Wrapper, and Seal is stored as $T_{bbox}$, $W_{cbbox}$, and $S_{cmp}$, respectively.
    As a whole system, $f^{recog}$ recognizes the current object using $img$ and $pc$ and reflects this in Target, Wrapper, and Seal information.
    Based on this, $f^{ctrl}$ provides trajectory commands to the real robot, and $f^{s/f}$ judges whether the operation succeeded or failed. If the result is a success, the process proceeds to the next step, repeating the cycle. If it fails, the process restarts from the initial state. At the end of all operations, $r$ is calculated by $f^{eval}$ to evaluate the final wrapping state.
  }
  \label{figure:whole-system}
\end{figure*}


\myswitchlang
  {ここではロボットによる包装動作を行ううえで、扱う物体や実験環境の仮定について述べたあと、それぞれの対象物体の表現方法について整理し、システム全体の概要を説明する。一般的に包装は包装される対象物、それを包み込むもの、そして包み込むものを封じるものの3つを主に使って行う。本論文ではこの3つをそれぞれTarget (T)、Wrapper (W)、Seal (S)と記号化する。}
  {In this section, we describe the assumptions made regarding the objects used in the wrapping operation and the experimental environment, organize how each object is represented, and give an overview of the overall system in the wrapping operation by the robot. Generally, wrapping involves using three key objects: the object to be wrapped, the object that wraps the object, and the object that seals the object. In this paper, these three objects are represented as Target (T), Wrapper (W), and Seal (S), respectively.}

\subsection{Assumptions of the Experimental Environment}
\myswitchlang
    {本実験では、扱う物体や実験環境を以下のように設定する。
      \begin{itemize}
        \item Targetは剛体であり、動作中に形状が変化しない。Target表面の凹凸や曲線形状は物体によって異なるが、それらを一般化し、おおまかに直方体と見做す。
        \item Wrapperは予めTargetを包み込むのに適した大きさの長方形状に切られているものとする。
        \item Sealは予め適切な長さに切られているものとする。
        \item 初期状態として、TargetはWrapperのおおよそ中央に傾きがなく平行に置かれているものとする。
      \end{itemize}
    }
    {In this experiment, the objects to be handled and the experimental environment are set up as follows:
      \begin{itemize}
      \item Target is rigid and does not change its shape during operation. Target is generalized to a rectangular body.
      \item Wrapper shall be a rectangular shape, pre-cut to a size suitable for enclosing Target.
      \item Seal shall be pre-cut to the appropriate length.
        \item Initially, Target is assumed to be placed in the approximate center of Wrapper with no tilt.
      \end{itemize}
    }

\subsection{Representation of the Object}

\subsubsection{Target}
\myswitchlang
    {
      本研究では直方体近似したTargetを、中心点の位置姿勢$T_{pose}$、寸法$T_{size}$の情報を保持するbounding box (bbox)($T_{bbox}$)として表現する。
      $T_{pose}$、$T_{size}$の情報からbboxの8頂点の位置が算出されるため、$T_{bbox}$は辺や面の位置姿勢情報も内包している。
      \figref{whole-system}の\``Additional Info for Target''に示すように、包装動作においては、WrapperでTargetを覆った後、その状態を保持しおさえ、もう片方のハンドでWrapperの端にリーチングしSealで固定する。そのため、TargetとWrapperを上からおさえる面$T_{hold}$とSealを貼る面$T_{seal}$の情報を$T_{bbox}$に加えてTargetの情報として保持することとする。$T_{hold}$について、Targetの形状によってTargetが転がったり滑ったりしないようにハンドでおさえる領域はある程度決まる。本研究では予め$T_{hold}$がTargetの適当な位置になるようTargetとWrapperが動作初期に配置されるものとする。$T_{seal}$はWrapperのエッジ部分が位置するTarget上の面となる。Sealを面に沿わせて貼り付けるために、Wrapperのエッジ位置($T_{seal-pos}$)と、その位置におけるTarget表面上の法線 ($T_{seal-\vec{n}}$)で$T_{seal}$を表現する。
      また、すべての包装動作の終了時に、Targetの表面に対してWrapperがどのように覆われたかを評価するため、Targetの点群を動作の開始時と終了時で比較する。包装動作の開始時と終了時のTargetの点群情報をそれぞれ$T_{pc-pre}$, $T_{pc-post}$と表す。
    }
    {
In this study, Target, approximated as a rectangular parallelepiped, is represented as a bounding box (bbox) ($T_{bbox}$), which contains the information of the central point's position and orientation ($T_{pose}$) and dimensions ($T_{size}$). Since the position of the eight vertices of the bbox is calculated from the information in $T_{pose}$ and $T_{size}$, $T_{bbox}$ also encapsulates the positional and orientational information of the edges and faces. As shown in the ``Additional Info for Target'' in \figref{whole-system}, during the wrapping operation, Target is held down by Wrapper after covering it, and one hand reaches the edge of Wrapper to secure it with Seal.
      Therefore, the information of the face used to hold Target ($T_{hold}$) and the face where Seal is applied ($T_{seal}$) is added to $T_{bbox}$.
      Regarding $T_{hold}$, the area held by the hand to prevent Target from rolling or slipping is determined to a certain extent depending on the shape of Target.
      In this study, we assume that Target and Wrapper are arranged at the beginning of the operation so that $T_{hold}$ is in an appropriate position for Target.
      $T_{seal}$ is the surface of Target where the edge of Wrapper is located. To apply Seal along the surface, $T_{seal}$ is expressed by the position of Wrapper's edge ($T_{seal-pos}$) and the normal vector of Target's surface at that position ($T_{seal-\vec{n}}$).
      Additionally, to evaluate how Wrapper covers the surface of Target after the wrapping process is completed, we compare the point clouds of Target before and after the operation. The point clouds at the start and end of the wrapping operation are denoted as $T_{pc-pre}$ and $T_{pc-post}$, respectively.
    }

\subsubsection{Wrapper}
\myswitchlang
    {

      Wrapperは一般的に紙や布、プラスチックフィルムなどTargetを包み込める柔らかい素材を用いる。これらは薄い柔軟物体であり、包装動作中に曲げたり折ったりすることで形状が変化するため、高さのある形状が一定なbboxによる状態の記述は難しい。
      そこで、まず、初期状態のWrapperを長方形とみなし、中心点の位置と姿勢 ($W_{pose}$)、縦横の長さ ($W_{size}$)、色 ($W_{color}$)の情報を格納した情報を本研究では高さがない色付きbounding box (cbbox)とし、これをWrapperの基本情報とする ($W_{cbbox}$)。
      $W_{pose}$、$W_{size}$の情報からcbboxの4頂点の位置が算出されるため、$W_{cbbox}$は辺の位置姿勢情報も内包している。この4辺や動作途中に現れる折り目の線などのエッジを用いてWrapperの状態を記述する。
      また、Wrapperの形状が初期の長方形から変化していくに伴い、色や注目領域を指定して抽出したWrapperの点群情報も用いることとなる ($W_{pc}$)。
   }
    {

      Wrapper is typically made of soft materials such as paper, cloth, or plastic film that can wrap Target.
      These are thin, flexible objects, and their shape changes when they are bent or folded during the wrapping process, so it is difficult to describe their state using a bbox with a fixed, tall shape.
      Therefore, in this study, we first consider the initial state Wrapper to be a rectangle, and store information on the position and posture of the center point ($W_{pose}$), the length of the width and height ($W_{size}$), and the color ($W_{color}$) as a colored bounding box (cbbox) with no height, which we use as the basic information for Wrapper ($W_{cbbox}$).
      Because the positions of the four vertices of the cbbox are calculated from the information in $W_{pose}$ and $W_{size}$, $W_{cbbox}$ also contains information about the positions and orientations of the edges. The state of Wrapper is described using these four edges and the edges of the creases that appear during the operation.

      Furthermore, as the shape of Wrapper changes from the initial rectangle, point cloud information of Wrapper, extracted by specifying $W_{color}$ and regions of interest, is also utilized ($W_{pc}$).
    }

\subsubsection{Seal}
\myswitchlang
    {
            Sealは一般的に紙やプラスチックでできたテープやシールが使われる。Sealは片面が粘着するという性質がある。また、丸められたりまとめられた状態から,ちぎったり破いたりして特定の大きさにしたりするため薄く柔らかく、加えて面積が小さい分張力も小さくなり、同様の素材でできたWrapperよりもさらに形状が不定となる。そのため、本研究では色 ($S_{color}$)と点群の平均XYZ座標 ($S_{mp}$)の情報を保持した情報 ($S_{cmp}$)でSealの状態を表すこととする。また、Wrapperと同様に色や注目領域を指定して抽出したSealの部分的な点群情報も用いることとなる ($S_{pc}$)
    }
    {
      Seal is typically made from tape or stickers composed of paper or plastic. A distinguishing feature of Seal is that one side is adhesive. Furthermore, they can be rolled or bundled, allowing them to be torn or ripped to specific sizes. Due to their thin and soft nature, as well as their smaller surface area, they exhibit lower tensile strength, resulting in a more indefinite shape compared to Wrapper made from similar materials. For this reason, in this study, the state of Seal is represented by information that retains color ($S_{color}$) and the mean XYZ coordinates of the point cloud ($S_{mp}$), referred to as $S_{cmp}$. Additionally, as with Wrapper, partial point cloud information of Seal, extracted by specifying color ($S_{color}$) and region of interest, is also utilized ($S_{pc}$).
    }


\subsection{Our Proposed System}
\myswitchlang
{本研究における包装を行うための全体のシステムを\figref{whole-system}に示す。包装動作を順にAからFまでの手順にわけ、それぞれの手順で
\begin{enumerate}
\item RGB画像 ($img$)と点群 ($pc$)によって現在の対象物体を認識し、Target, Wrapper, Sealの情報に反映させる ($f^{recog}$)
\item Target, Wrapper, Sealの情報をもとに実機ロボットに軌道指令を与える ($f^{ctrl}$)
\item 2の動作の成否を判定する ($f^{s/f}$)
\end{enumerate}
を行う。3で動作が成功と判定された場合は次の動作にうつり、失敗と判定された場合は初期状態からやり直す。ロボットから見てTargetの右側の面を覆う動作の各手順における$f^{recog}$、$f^{ctrl}$, $f^{s/f}$の入力と出力を明らかにし、一般化して表現したものを\tabref{systemtable}に示す。$f^{s/f}$はロボット又はTargetに対して柔軟物体であるWrapperとSealが適切に配置されているかを認識・評価するものである。注目領域 ($ROI$)と各物体の色による抽出点群 ($W_{pc}$又は$S_{pc}$)を入力とし、$ROI$内の抽出点群の有無を判定する関数で一般化できる。
また、最後の動作Fにおいて、$f_{eval}$関数によって$r$の値を算出することによって最終的な包装状態を評価する。$r$については次章Fで説明する。
}
{This study's overall wrapping system is shown in \figref{whole-system}. The wrapping process is divided into six sequential steps, A through F. Each step involves the following procedures:
\begin{enumerate}
\item $f^{recog}$: Recognize the current target object by using RGB image ($img$) and point cloud ($pc$) and reflect it in the information of Target, Wrapper, and Seal.
\item $f^{ctrl}$: Send trajectory commands to the actual robot based on the information in Target, Wrapper, Seal.
\item $f^{s/f}$: Judge the success or failure of the motion of 2.
\end{enumerate}
The inputs and outputs of $f^{recog}$, $f^{ctrl}$, and $f^{s/f}$ in each step of the operation to cover the right side of Target as seen from the robot are clarified and expressed in generalized form in \tabref{systemtable}. The $f^{s/f}$ recognizes and evaluates whether the flexible objects, Wrapper and Seal, are appropriately placed relative to the robot or Target. It can be generalized by a function that takes a region of interest ($ROI$) and a set of extracted points ($W_{pc}$ or $S_{pc}$) by the color of each object as input and determines the presence or absence of the extracted point set in $ROI$.
Additionally, in the final operation F, the final wrapping state is evaluated by calculating the value of $r$ using the $f^{eval}$ function. The value of $r$ is explained in the next chapter F.
}

\myswitchlang
{次章以降でこれらのシステムの各手順の具体的な関数の内容を説明し実験を通して本章で提案した各対象物体の表現や関数の妥当性を示す。座標系や$T_{bbox}$, $W_{cbbox}$における各頂点のロボットに対する位置を\figref{whole-system}示し、例えばTargetの面abcdやWrapperの辺abは$T_{abcd}$, $W_{ab}$のように示す。}
{In the following chapters, we explain the specific functions of each procedure of these systems and show the validity of the representations and functions of each object proposed in this chapter through experiments. The coordinate system and the position of each vertex in $T_{bbox}$ and $W_{cbbox}$ relative to the robot are shown in \figref{whole-system}. For example, the face abcd of Target and the edge ab of Wrapper are shown as $T_{abcd}$ and $W_{ab}$.}

\begin{table*}[t]
  \caption{Procedure for wrapping the right side of the Target, including the necessary hardware design requirements, recognition, and control/manipulation methods for both left and right hands}
  \label{table:systemtable}
  \begin{adjustbox}{width=\textwidth}
    \begin{tabular}{c|c|l|l|l|l} \hline
      & Procedure & Hardware & Recognition & Rhand-Manipulation & Lhand-Manipulation \\ \hline \hline
      \multirow{4}{*}{A} & \multirow{4}{*}{Preoperative Recognition} & & $T_{bbox} \gets f^{recog}_{1}(pc)$ & & \\
      & & & $W_{cbbox} \gets f^{recog}_{2}(img, pc)$ & & \\
      & & & $S_{cmp} \gets f^{recog}_{3}(pc)$ & & \\
      & & & $T_{pc-pre} \gets f^{recog}_{4}(pc, T_{bbox}, W_{cbbox})$ & & \\ \hline
      B & Lifting Wrapper Upward & Hardness and Thinness of Nails & $s/f \gets f^{s/f}(ROI = rhand coords, W_{pc})$ & $f^{ctrl}_{2}(edge = W_{ab})$ & $f^{ctrl}_{1}(\vec{F} = (0, +, 0), face = T_{cdhg})$ \\ \hline
      \multirow{3}{*}{C} & \multirow{3}{*}{Covering} & \multirow{3}{*}{
        \begin{tabular}{c}
          Three-dimensional Shape \\ and Elasticity of the Fingers
        \end{tabular}}
      & \multirow{3}{*}{$s/f \gets f^{s/f}(ROI = T_{abcd-}, W_{pc})$} & $f^{ctrl}_{3}(edge = T_{ef}, \vec{F} = (0, 0, +f), collision = False)$ & \multirow{3}{*}{$\downarrow$} \\
      & & & & $f^{ctrl}_{3}(edge = T_{ab}, \vec{F}(0, +f, 0), collision = True)$ & \\
      & & & & $f^{ctrl}_{4}(constraint = Target)$ & \\ \hline
      D & Gripping Seal & & $s/f \gets f^{s/f}(ROI = lhand coords, S_{pc})$ & & $f^{ctrl}_{5}(S_{mp})$ \\ \hline
      \multirow{4}{*}{E} & \multirow{4}{*}{Securing with Seal}
      &
      \multirow{4}{*}{
      \begin{tabular}{c}
        Three-dimensional Shape \\ and Elasticity of the Fingers
        \end{tabular}}
        &
      $T_{seal-pos} \gets f^{recog}_{5}(img, pc, T_{bbox}, W_{cbbox})$ & & \multirow{4}{*}{$f^{ctrl}_{4}(constraint=Seal)$} \\
      & & & $T_{seal-\vec{n}} \gets f^{recog}_{6}(pc, T_{seal-pos}, thre)$ & & \\
      & & & $s/f \gets f^{s/f}(ROI = T_{abcd}, S_{pc})$ & & \\
      & & & $s/f \gets f^{s/f}(ROI = T_{abcd-}, W_{pc})$ & & \\ \hline
      \multirow{3}{*}{F} & \multirow{3}{*}{
      \begin{tabular}{c}
        Evaluation of \\ Wrapping Condition
      \end{tabular}}
      & & $T_{pose}, W_{pose} \gets f^{recog}_{5}(img, pc, T_{bbox}, W_{cbbox})$ & & \\
      & & & $T_{pc-post} \gets f^{recog}_{4}(pc, T_{bbox}, W_{cbbox})$ & & \\
      & & & $r \gets f^{eval}(T_{pc-pre}, T_{pc-post}, k)$  & & \\ \hline
    \end{tabular}
  \end{adjustbox}
\end{table*}

\section{Integrative wrapping system for dual-arm humanoid}
\myswitchlang
{ここでは、前章で示した全体のシステムについて、各要素のハードウェア、認識、左右ハンドの制御方法の詳細を説明する。}
{This section details the hardware, recognition, and left and right-hand control methods for each system element presented in the previous chapter.}

\subsection{Preoperative Recognition}
\myswitchlang
{ まず最初にTargetとWrapper、Sealの初期情報を認識する。$T_{bbox}$は$pc$を用いて物体が置かれている平面上の点群を抽出し境界ボックスを認識して求める ($f^{recog}_{1}$)。Wrapperは薄いため物体が置かれている平面上の点群からbboxを抽出することが難しい。そのため、$img$を用いたエッジ検出によってピクセル座標上の4頂点を認識した後、$pc$を用いて三次元座標に変換し、$W_{cbbox}$を求める ($f^{recog}_{2}$)。$S_{cmp}$は$pc$を色抽出しその点群の平均XYZ位置座標して求める ($f^{recog}_{3}$)。このようにして得られた$T_{bbox}$と$W_{cbbox}$の情報をもとに、Targetを直方体近似したときにWrapperにTargetが覆われる面を概算し$T_{pc-pre}$を求める ($f^{recog}_{4}$)。}
{First, the initial information of the Target, Wrapper, and Seal is recognized. The $T_{bbox}$ is obtained by extracting a set of points on the plane where the object is placed using $pc$ and recognizing the bounding box ($f^{recog}_{1}$). Because the Wrapper is thin, extracting the bbox from the point cloud on the plane where the object is placed is difficult. Therefore, after recognizing the four vertices in pixel coordinates by edge detection using $img$, $W_{cbbox}$ is obtained by converting it to 3D coordinates using $pc$ ($f^{recog}_{2}$). The $S_{cmp}$ is obtained as the average XYZ position of the point cloud by color extraction of $pc$ ($f^{recog}_{3}$). Based on the information of $T_{bbox}$ and $W_{cbbox}$ obtained in this way, the surface where the Target is covered by the Wrapper when a rectangle approximates the Target is estimated, and $T_{pc-pre}$ is obtained ($f^{recog}_{4}$).}

\subsection{Lifting Wrapper Upward}
\begin{figure}[thbp]
  \centering
  \includegraphics[width=0.25\columnwidth]{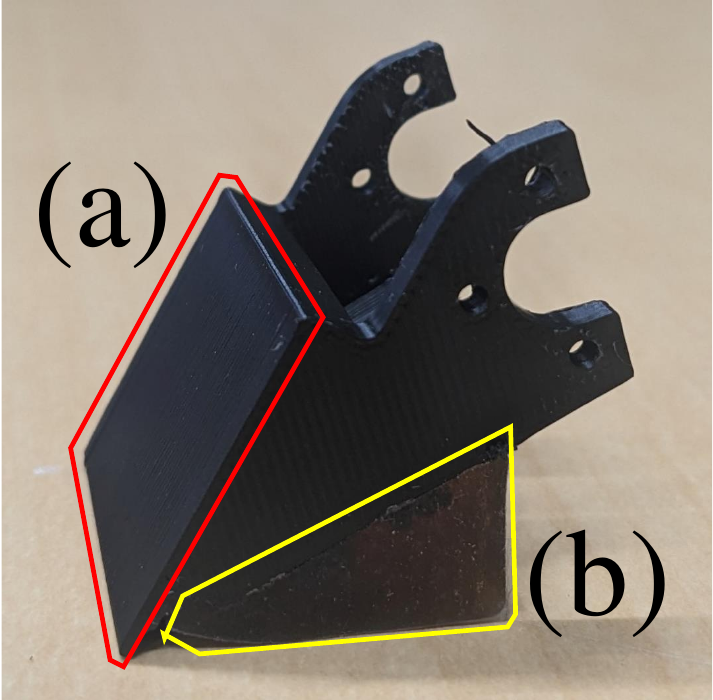}
  \caption{(a) realizes the thin, hard, nail-shaped part required in step B. (b) realizes the three-dimensionality and elasticity required in steps C and E by means of rubber molding.}
  \label{figure:hand-pic}
\end{figure}

\begin{figure}[thbp]
  \centering
  \includegraphics[width=0.9\columnwidth]{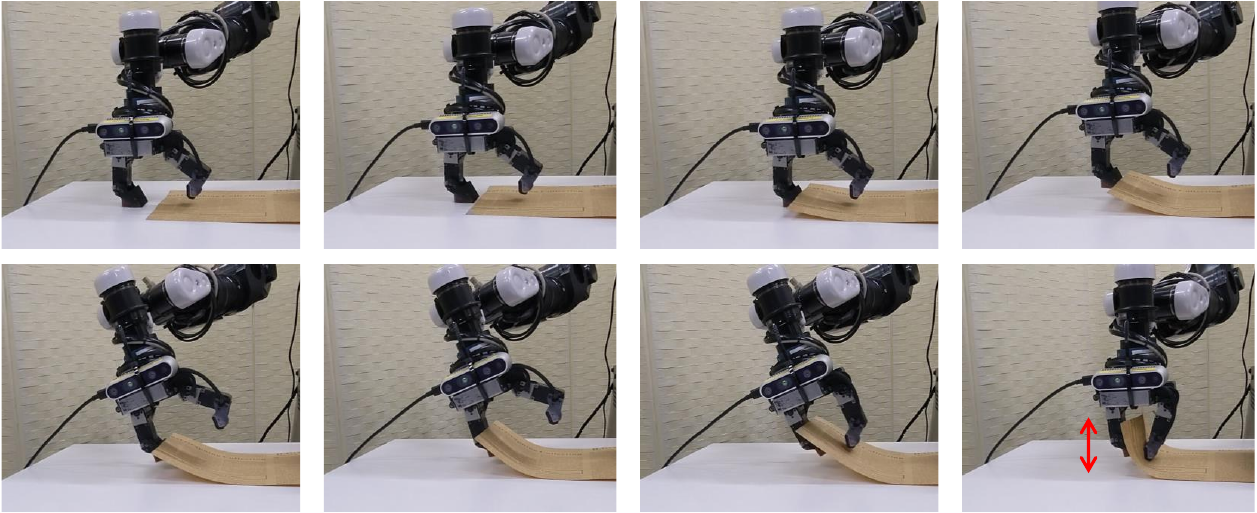}
  \caption{Lifting Wrapper from the plane; after grasping Wrapper, the hand is tilted and moved in the direction of Wrapper's center to secure enough $T_{seal}$ area (red arrow).
}
  \label{figure:turn-over}
\end{figure}

\begin{figure}[thbp]
  \centering
  \includegraphics[width=0.9\columnwidth]{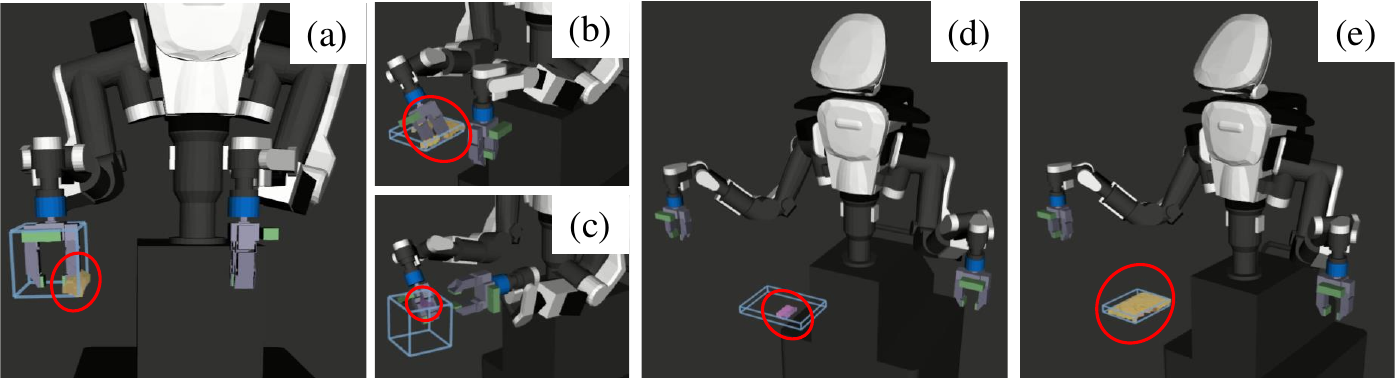}
  \caption{The success or failure of each operation is determined by the presence or absence of the extracted points ($W_{pc}$ or $S_{pc}$) in the $ROI$ ($f^{s/f}$). A blue box represents the $ROI$.}
  \label{figure:sf}
\end{figure}

\myswitchlang
{Wrapperを平面から持ち上げる。まず動作中にTarget, Wrapperの位置がずれたり滑ったりしないよう、左ハンドでおさえる。ハンドがWrapperを引っ張る向き (\vec{F})にTargetが動いてしまうのを妨げる反力をかけるTarget面 ($face$)にハンドをリーチングする ($f^{ctrl}_{1}$)。ロボットから見てTargetの右側を覆う場合、Wrapperをy軸正方向 ($\vec{F} = (0, +, 0)$)に力をかけて引っ張るため、$T_{cdhg}$にハンドが接するようにリーチングする。次に右ハンドでWrapperの$edge$にリーチングし、Wrapperを持ち上げる。このときWrapperとWrapperが置かれた平面上の間にハンドを差し込んで把持するために、薄く硬い爪のような形状を有したハンド (\figref{hand-pic}のA部分)でWrapperをすくいあげる ($f^{ctrl}_{2}$)。動作の様子は\figref{turn-over}で示される。もう片方のハンドで行うSealを貼る動作Eを行えるよう、$T_{seal}$領域を確保するため、Wrapperを平面から持ち上げた後ハンドを傾け、ハンドをWrapperの中心方向へ動かした後、Wrapperをつかむ。動作の成否判定は、把持動作終了時の右ハンドの座標を中心にとる領域内に$W_{pc}$が存在するかどうかによって判定する (\figref{sf}の(a))。}
{Lift the Wrapper from the plane. Reaching the hand to the Target surface ($face$), which exerts a reaction force that prevents the Target from moving in the direction that pulls the Wrapper ($f^{ ctrl}_{1}$). To cover the right side of the Target, as seen by the robot, the Wrapper is pulled with force in the positive direction of the y-axis ($\vec{F} = (0, +, 0)$), so the hand is reached so that it touches $T_{cdhg}$. Next, the right hand reaches the $edge$ of the Wrapper and lifts the Wrapper. At this time, in order to insert the hand between the Wrapper and the plane on which the Wrapper is placed and grasp it, the hand ((a) of \figref{hand-pic}), which has a thin and hard nail-like shape, scoops up the Wrapper ($f^{ctrl}_{2}$). The action is shown in \figref{turn-over}. In order to secure the $T_{seal}$ area for the operation E, which is performed with the other hand, the hand is tilted after lifting the Wrapper from the plane, and the hand is moved toward the center of the Wrapper, and then the Wrapper is grasped. The success or failure of the operation is judged by whether $W_{pc}$ exists in the region centered at the coordinates of the right hand at the end of the grasping operation as shown in (a) of \figref{sf}.}

\subsection{Covering}
\begin{figure}[thbp]
  \centering
  \includegraphics[width=\columnwidth]{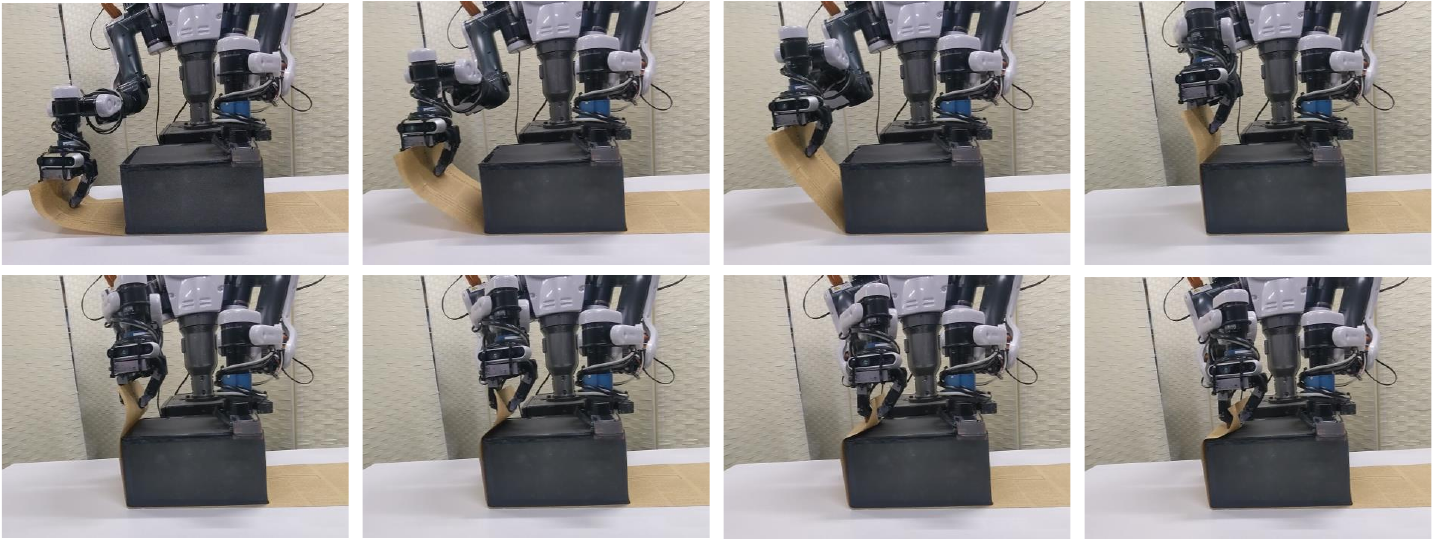}
  \caption{Cover one side of Target while tensioning the paper using admittance control. The upper row is the operation of covering $T_{abfe}$ and the lower row is the operation of covering $T_{abcd}$.}
  \label{figure:cover}
\end{figure}

\myswitchlang
{把持したWrapperをテンションを張りながらTargetの面を覆う。この動作を$f^{ctrl}_{3}(edge, \vec{F}, collision)$として$T_{bbox}$の一面を覆う動作を一般化し、アドミッタンス制御を用いて一定の力でWrapperを引っ張りながらハンドの位置座標を端点に円弧状にハンドを動かす。関数の引数である$edge$は円弧の中心点の位置する$T_{bbox}$の辺、$\vec{F}$はアドミッタンス制御によるWrapperを引っ張る力の向き、$collision$は動作修了時のTargetとハンドとの衝突判定を行うかどうかを表す。\figref{cover}の上段は$T_{abfe}$を覆う動作であり、$T_{ef}$の中点を円弧の中心点とし、z軸正方向にf [N]引っ張る力をかけた状態で動作させる。この場合は動作終了時にハンドは空中に位置しTargetの面との衝突は起こらない。このような場合引数$collision$は$False$とし、90度の円弧を描くように動作させる。\figref{cover}の下段は$T_{abcd}$を覆う動作であり、$T_{ab}$の中点を円弧の中心点とし、y軸正方向にf [N]引っ張る力をかける。動作終了時はハンドとTargetが衝突する。このような場合引数$collision$は$True$とし、力覚によって衝突を検知したとき動作を覆う動作を中断する。}
{The grasped Wrapper is covered with the Target's face while tensioning it. This motion is generalized as $f^{ctrl}_{3}(edge, \vec{F}, collision)$ to cover one face of $T_{bbox}$, and the hand is moved in an arc with the hand position coordinates as the endpoints while pulling the Wrapper with constant force using admittance control. The function argument $edge$ is the edge of $T_{bbox}$ where the center point of the arc is located, $\vec{F}$ is the direction of the force pulling the Wrapper by admittance control, and $collision$ indicates whether or not the collision between Target and the hand is judged after the operation. The upper row of \figref{cover} is the operation to cover $T_{abfe}$, with the midpoint of $T_{ef}$ as the center point of the arc and the force to pull f [N] in the z-axis positive direction is applied. In this case, the hand is positioned in the air at the end of the movement, and there is no collision with the Target's surface. In such a case, the argument $collision$ is set to $False$, and the hand moves in a 90-degree arc. The lower row of \figref{cover} is a move to cover $T_{abcd}$. The middle point of $T_{ab}$ is the arc's center point, and the force to pull f [N] in the y-axis positive direction is applied. At the end of the operation, the hand and Target collide. In such a case, the argument $collision$ is set to $True$, and when the force sensor detects the collision, the motion is aborted.}

\begin{figure}[thbp]
  \centering
  \includegraphics[width=\columnwidth]{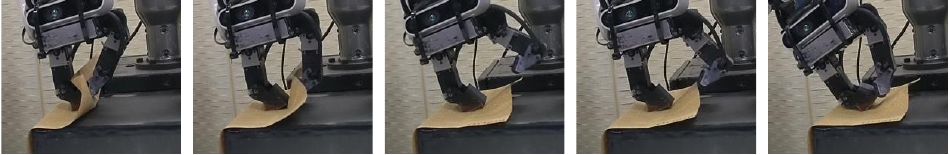}
  \caption{Motion of pulling out one finger while keeping the Wrapper pressed against the hand using the three-dimensionality and elasticity of the hand.}
  \label{figure:pull-finger}
\end{figure}

\myswitchlang
{この後、Wrapperを2指で把持した状態からWrapperの下に潜り込んだ1指を離す。このとき、以下のことを同時に満たしながら連続的に動作を行う必要がある。
\begin{itemize}
\item 紙の下に潜り込んだ1指を離す
\item テンションが張った状態のWrapperの状態を保つ
\item $T_{seal}$領域を十分に確保する
\end{itemize}
以上の条件を満たすように、立体性とやわらかさを有するハンドによって$T_{hold}$領域を押し付けながらハンドを回転させて潜り込んだ指を抜く ($f^{ctrl}_{4}$)。本実験ではWrapperを把持したときに外側に位置する方の指の背にゴムを成形しこの条件を満たす (\figref{hand-pic}の(b))。動作の様子を\figref{pull-finger}で示す。この関数の入力$constraint$はこの動作では$Target$である。$constraint$引数についてはEで詳しく説明する。}
{After this, release one finger that has gone under the Wrapper. At this time, the operation must be performed continuously while simultaneously satisfying the following:
\begin{itemize}
\item Release one finger under the paper
\item Keep the Wrapper in a tensioned state
\item Allocate sufficient $T_{seal}$ area
\end{itemize}
In order to satisfy the above conditions, the hand is rotated, and the hidden finger is removed while pressing down on the $T_{hold}$ area by the hand having three-dimensionality and elasticity ($f^{ctrl}_{4}$). In this experiment, this condition is satisfied by molding rubber on the back of the finger that is positioned on the outside when the Wrapper is grasped as shown in (b) of \figref{hand-pic}. The operation is shown in \figref{pull-finger}. The input $constraint$ of this function is $Target$ in this operation. The $constraint$ argument is described in detail in E.}

\myswitchlang
{動作の成否判定は、$T_{abcd}$のロボットに対して右半平面の領域内 ($T_{abcd-}$)に$W_{pc}$が存在するかどうかによって判定する (\figref{sf}の(b))。動作が成功した場合は点群が現れるが、途中で紙が滑り落ちるなど把持に失敗した場合は現れない。}
{The success or failure of the motion is judged by whether or not $W_{pc}$ exists in the right half-plane region in $T_{abcd}$  as seen
from the robot ($T_{abcd-}$)  as shown in (b) of \figref{sf}. Point clouds appears if the motion is successful, but if the grasp fails because the paper slips down in the process, they do not appear.}

\subsection{Gripping Seal}
\myswitchlang
    {Sealを把持する。$S_{mp}$を入力にSealにリーチした後、Sealを剥がすようにハンドを動かす ($f^{ctrl}_{5}$) (\figref{grip-tape})。動作の成否判定は左ハンド前方の領域内に$S_{pc}$が存在するかどうかによって判定する (\figref{sf}の(c))。把持に成功した場合は左ハンドに追従して$S_{pc}$が現れるはずである。}
{The Seal is grasped. $S_{mp}$ is used as input. After reaching for the Seal, the hand is moved as if peeling off the Seal ($f^{ctrl}_{5}$). The success or failure of the operation is judged by whether $S_{pc}$ exists in the area in front of the left hand as shown in (c) of \figref{sf}. If the grasp is successful, $S_{pc}$ should appear following the left hand.}

\subsection{Securing with Seal}
\begin{figure}[thbp]
  \centering
  \includegraphics[width=0.85\columnwidth]{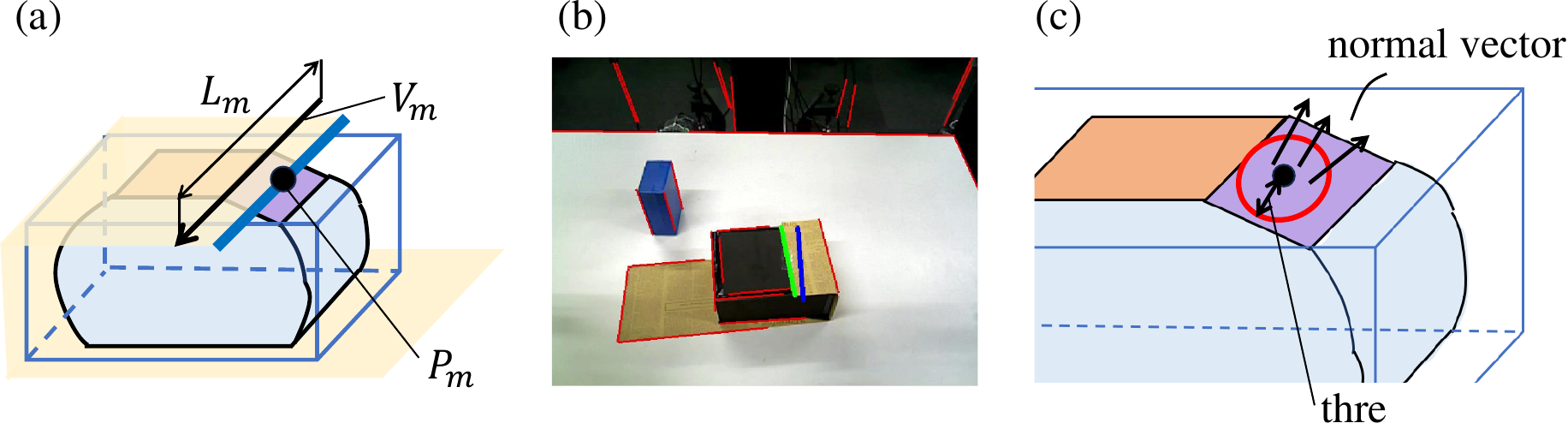}
  \caption{Find the position of the edge of the Wrapper using $T_{bbox}$ and $W_{cbbox}$ (a), and calculate the similarity using edge detection to find the actual edge (b). $T_{seal-\vec{n}}$ is the average of the normal vectors around $T_{seal-pos}$ (c).}
  \label{figure:secure-recog}
\end{figure}

\myswitchlang
    {SealをWrapperに貼り封をしてTargetに固定する。$T_{seal-pos}$、$T_{seal-\vec{n}}$を認識し、それに基づき左ハンドでリーチングする。$T_{seal-pos}$は以下のようにして求める。まず$img$を用いてエッジ検出をした後 (\figref{secure-recog}の(b)で示す赤い線)、$pc$を用いて各エッジの端点のピクセル座標を三次元座標に変換し、各エッジの中心点の座標$P_i$, 長さ$L_i$, 方向ベクトル$V_i$を求める。
      一方で$T_{bbox}$, $W_{cbbox}$から算出される、SealをはるWrapperの辺について、中心点の座標$P_m$, エッジの長さ$L_m$,エッジの方向ベクトル$V_m$を求める (\figref{secure-recog}の(a))。
      それと各エッジとの差$D_{P_i}=||P_i - P_m||$, $D_{L_i}=|L_i - L_m|$, $D_{V_i}=1-abs(\frac{|V_i\cdot V_m|}{||V_i||||V_m||})$を重み付けして合計したものを類似度として評価する。
\begin{equation}
\label{eq}
  S_i = w_PD_{P_i} + w_LD_{L_i} + w_VD_{V_i}
\end{equation}
類似度$S_i$が小さいものを実際のTargetの形状に即してWrapperが覆われたときのWrapperのエッジとする(緑の線)。このようにして実際のTargetの形状に即し、これまでの手順による対象物体の移動やずれに対応した正確な$T_{seal-pos}$を求める ($f^{recog}_{5}$)。$T_{seal-\vec{n}}$は$pc$のうち、$T_{seal-pos}$との距離が閾値 ($thre$)以下の点における、Targetの面における法線の平均とする ((c))($f^{recog}_{6}$)。}
    {The Seal is attached to the Wrapper and fixed to the Target. Recognize $T_{seal-pos}$ and $T_{seal-\vec{n}}$ and reach with the left hand based on them. The $T_{seal-pos}$ is obtained as follows. First, after edge detection using $img$ (red line shown in (b) of \figref{secure-recog}), the pixel coordinates of the endpoints of each edge are converted to three-dimensional coordinates using $pc$ to obtain the coordinates of the center $P_i$, the edge length $L_i$, and the edge direction vector $V_i$  of each edge.
      On the other hand, for the edge of the Wrapper where the Seal is applied, calculated from $T_{bbox}$ and $W_{cbbox}$, the coordinates of the center $P_m$, the edge length $L_m$, and the edge direction vector $V_m$ are determined ((a) of \figref{secure-recog}).
      The difference between it and each edge $D_{P_i}=||P_i - P_m||$, $D_{L_i}=|L_i - L_m|$, $D_{V_i}=1-abs(\frac{|V_i\cdot V_m|}{||V_i||||V_m||})$ is weighted and summed to evaluate as similarity.
\begin{gather}
  S_i = w_PD_{P_i} + w_LD_{L_i} + w_VD_{V_i}
\end{gather}
The one with the smallest similarity $S_i$ is estimated as the edge of the Wrapper when the Wrapper is covered with the actual Target shape (green line). In this way, the exact $T_{seal-pos}$ is obtained ($f^{recog}_{5}$), which corresponds to the actual shape of the Target and the movement and displacement of the target object by the previous procedure.
$T_{seal-\vec{n}}$ is defined as the average of the normals on the Target surface at the points in $pc$ that are within the threshold ($thre$) distance from $T_{seal-pos}$ ($f^{recog}_{6}$) as shown in (c) of \figref{secure-recog}.
    }

\begin{figure}[thbp]
  \centering
  \includegraphics[width=0.9\columnwidth]{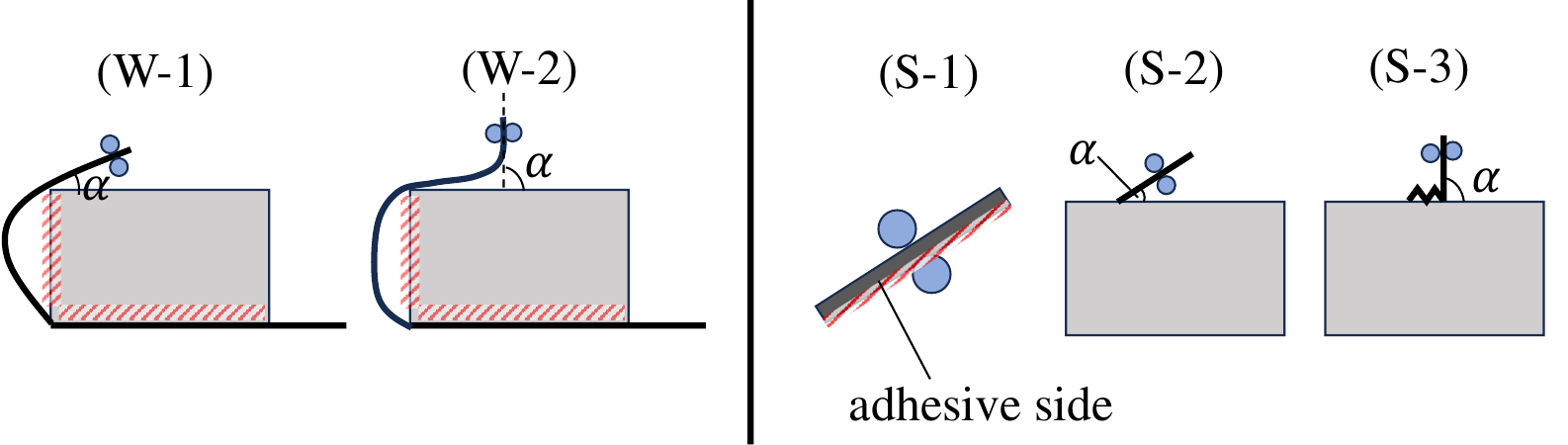}
  \caption{Difference of $constraint$ in $f^{ctrl}_{4}$. In step C, the Wrapper is bound to the surface of the Target and is restrained on one end, so the tension of the Wrapper can be maintained even if the angle $\alpha$ is not small (W-2). On the other hand, in step E, the restraint is on the adhesive side (S-1), and both ends of the Seal are free, so if the value of $\alpha$ is large, as in (S-3), the Seal will be deformed. Therefore, it reaches Target with a small $\alpha$ on the surface, as shown in (S-2).}
  \label{figure:tape-secure-constraint}
\end{figure}

\begin{figure}[thbp]
  \centering
  \includegraphics[width=\columnwidth]{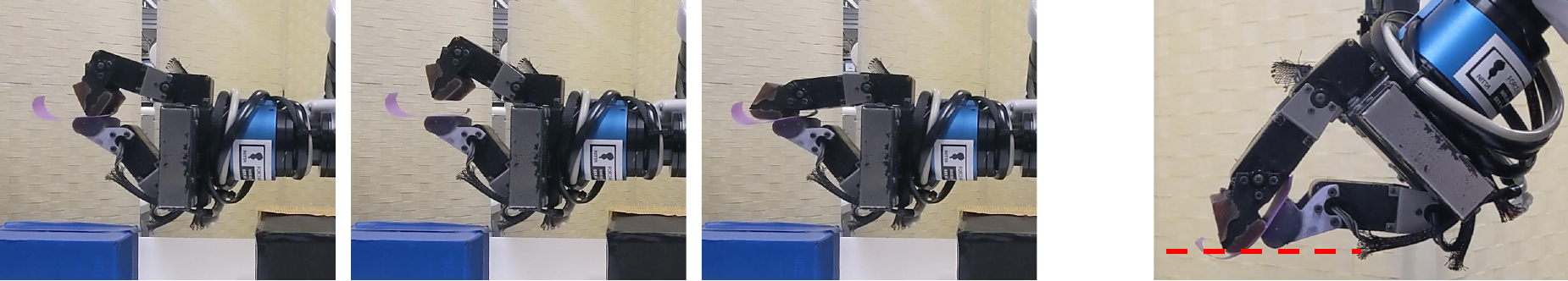}
  \caption{From the state where the fingers are relative to each other, move the fingers to a state where the upper fingers are not on the adhesive side of the Seal and approach the surface before the lower fingers on the adhesive side.}
  \label{figure:tape-secure-hand}
\end{figure}

\begin{figure}[thbp]
  \centering
  \includegraphics[width=\columnwidth]{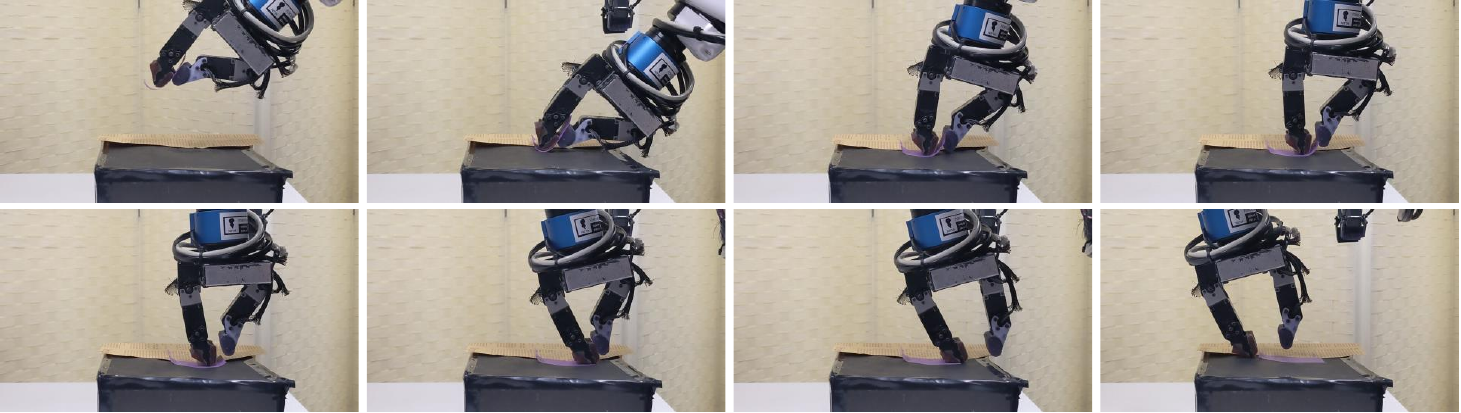}
  \caption{Operation to apply Seal. By pressing down on the non-adhesive side of the Seal, the fingers are released from the adhesive surface, and the Seal can be applied by pushing the lower fingers away from it and using the hand's three-dimensional shape and elasticity to press down on it.}
  \label{figure:tape-secure}
\end{figure}

\myswitchlang
    {以上で認識した$T_{seal-pos}$にハンドをリーチしてSealを貼る。動作としては手順Cと同様に2指で把持した状態から下に潜り込んだ1指を離すものである。しかし、手順Cでは把持対象であるWrapperはTargetの下に置かれ、側面を覆い込むことでTarget底面や側面に拘束されWrapperは片端拘束の状態となる。そのため\figref{tape-secure-constraint}の(W-1)のようにリーチするTargetの面との角度$\alpha$が小さくなくてもWrapperのテンションを保つことができる(W-2)。

      しかし、手順Eでは拘束は粘着面にあり(S-1)、Sealの両端は自由となっているため、(S-3)のように$\alpha$の値が大きいとSealはよれてしまう。そのため、(S-2)で示すように面に$\alpha$が小さい状態でTargetにリーチしなければならない。このとき、\figref{tape-secure-hand}に示すように、手順DでSealを把持するために指が相対した状態から、Sealの粘着面側ではない上側の指が粘着面側の下側の指よりも先に面に近づく状態に指を移動させる。こうすることでSealの粘着面側ではない面をおさることで指と粘着面との拘束を解きながら下の指をSealから離し、ハンドの立体性とやわらかさを用いておしつけながらSealを貼ることができる ($f^{ctrl}_{4}$)(\figref{tape-secure})。}
    {Then, reach the hand to $T_{seal-pos}$ recognized by the above and affix the seal. The action is the same as in step C, where the hand is grasped with two fingers, and the finger that has gone under is released. However, in step C, the Wrapper is placed under the Target, and by covering the sides, the Wrapper is bound to the bottom and sides of the Target, and the Wrapper is bound to one end. Therefore, the Wrapper can be kept in tension even if the angle $\alpha$ with the face of the reaching Target is not small as in (W-1) and (W-2) of \figref{tape-secure-constraint}.

However, in step E, the constraint is on the adhesive surface (S-1), and both ends of the Seal are free, so if the value of $\alpha$ is large, as shown in (S-3), the Seal will deform. Therefore, it must reach Target with a small $\alpha$ on the surface as shown in (S-2). In this case, as shown in \figref{tape-secure-hand}, from the state where fingers are relative to grasp the Seal in step D, the upper fingers that are not on the adhesive side of the Seal are moved to approach the surface before the lower fingers on the adhesive side. In this way, the lower fingers are separated from the Seal while the restraint between the fingers and the adhesive side is released by pushing the non-adhesive side of the Seal back. The three-dimensionality and elasticity of the hand are used to apply the Seal while pressing down ($f^{ctrl}_{4}$) as shown in \figref{tape-secure}.}

\myswitchlang
{動作の成否判定はまず、$T_{abcd}$内に$S_{pc}$が存在するかどうかによってSealを貼り付けられたかどうかを判定する (\figref{sf}の(d))。次にSealによってWrapperが固定されたかどうかを、$T_{abcd-}$内に$S_{pc}$が存在するかどうかによって判定する。 (\figref{sf}の(e))}
{The success or failure of the operation is first determined by whether or not $S_{pc}$ exists in $T_{abcd}$ to determine whether or not the Seal has been pasted as shown in (d) of \figref{sf}. Next, determine if the Wrapper has been fixed by the Seal by whether $S_{pc}$ exists in $T_{abcd-}$ as shown in (e) of \figref{sf}.}


\subsection{Evaluation of Wrapping Condition}
\myswitchlang
    {
      包装状態を評価する。これまでの動作で初期状態から物体が移動した可能性があるため、$T_{pose}$と$W_{pose}$を更新する。手順Eの$T_{seal-pos}$を求めたときと同様、$f^{recog}_{5}$を用いて、TargetまたはWrapperの辺のどれかについて、エッジの類似度を用いて移動を推定し、$T_{pose}$と$W_{pose}$、ひいては$T_{bbox}$、$W_{cbbox}$を更新する。その更新された$T_{bbox}$、$W_{cbbox}$を用いて、$T_{pc-pre}$と同様に$T_{pc-post}$を取得する。$T_{pc-pre}$と$T_{pc-post}$をICPマッチングして位置合わせを行った後、以下のような式を用いて包装状態を評価する ($f^{eval}$)。

\begin{gather}
  Neighborhood(B_{i}) = \{a_{j1}, a_{j2}, \cdots , a_{jk}\} \\
  \overline{\vec{n}}_{Ai} = \frac{1}{k}\sum^{k}_{j=1}\vec{n_{Aij}} \\
  r = \frac{\sum^{m}_{i=1}1(acos(\frac{\overline{\vec{n}}_{Ai} \dot \vec{n}_{Bi}}{||\overline{\vec{n}}_{Ai}||||\vec{n}_{Bi}||}) \geq 10)}{M}
\end{gather}}
    {
      Evaluate the wrapping state. Since there is a possibility that the object has moved from the initial state during the previous operations, $T_{pose}$ and $W_{pose}$ should be updated. In the same way as when calculating $T_{seal-pos}$ in Step E, movement is estimated using the edge similarity for one of the Target or Wrapper edges with $f^{recog}_{5}$, and $T_{pose}$ and $W_{pose}$, and in turn $T_{bbox}$ and $W_{cbbox}$, are updated.
      Using the updated $T_{bbox}$ and $W_{cbbox}$, $T_{pc-post}$ is obtained in the same way as $T_{pc-pre}$.
      After aligning $T_{pc-pre}$ and $T_{pc-post}$ through ICP matching, the wrapping state is evaluated using the following equation ($f^{eval}$).
\begin{gather}
  \overline{\vec{n}_{Ai}} = \frac{1}{k}\sum^{k}_{j=1}\vec{n}_{Aij} \\
  r = \frac{\sum^{m}_{i=1}1(acos(\frac{\overline{\vec{n}_{Ai}} \cdot \vec{n}_{Bi}}{||\overline{\vec{n}}_{Ai}||||\vec{n}_{Bi}|}) \geq 10)}{ M}
\end{gather}}

\begin{figure}[thbp]
  \centering
  \includegraphics[width=0.8\columnwidth]{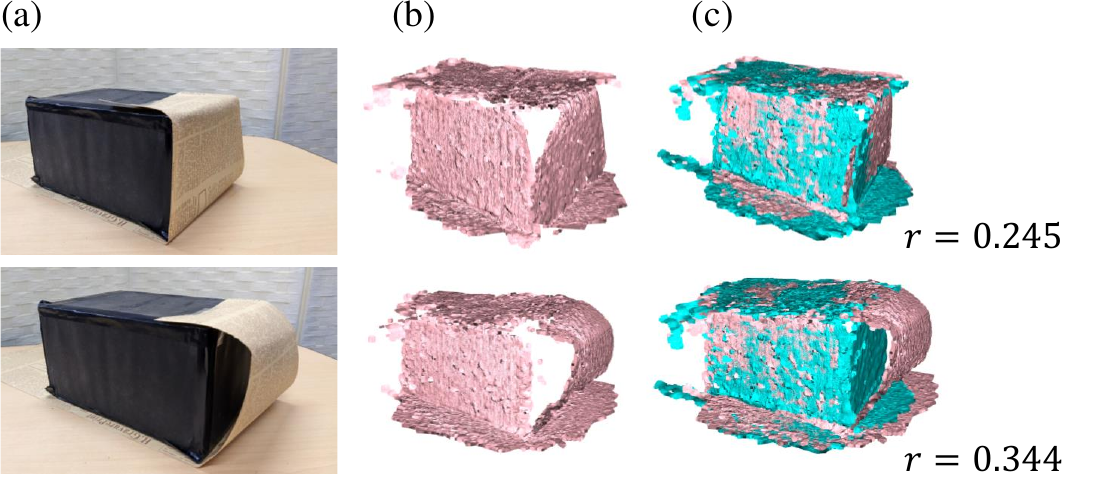}
  \caption{The top row represents a good example where the Wrapper follows the face of the Target, and the bottom row represents a bad example where the Wrapper bulges out. (b) shows $T_{pc-post}$ in (a) and (c) shows $T_{pc-pre}$ (blue) and $T_{pc-post}$ (red) icp-matched. Determine the state of wrapping by the size of the value of $r$, which is the percentage of normal vectors that are not along the face of the Target.}
  \label{figure:evaluation-compare}
\end{figure}

\myswitchlang
    {
      上の式で、$A$、$B$はそれぞれICPマッチングされた後の$T_{pc-pre}$、$T_{pc-post}$の点群を表す。また、$a$, $b$はそれぞれ$A$、$B$の各点の座標を、$\vec{n}_{Ai}$はAのある点$a_{i}$における法線ベクトルを表す。まずBの各点$b_{i}$に対して、その点と距離が近いものから順に$k$個の$A$の点群座標($a_{i1}$, $a_{i2}$, ...)を取り出し、$Neighborhood(B_{i})$とする。次に$Neighborhood(B_{i})$の各点における法線ベクトルを平均し、それを$\overline{\vec{n}_{Ai}}$とする。その後$B$の各点において、$\vec{n}_{Bi}$と$\overline{\vec{n}_{Ai}}$のなす角度を計算し、その角度が10度以上となる点の数を数え、$B$の集合の大きさ$M$で割る。これによって、$T_{pc-pre}$と$T_{pc-post}$を比較したときに各点における法線の状態が大きく異なるものの割合が算出される。つまり、包装動作終了時にWrapperがどの程度Targetの表面にしっかりと沿って包装されているかがわかる。\figref{evaluation-compare}の上段はTargetの面に沿って紙のテンションを保ち包装できた良い例を示す。下段は紙がふくらんでしまった悪い例を表す。$r$の値はそれぞれ0.245, 0.344となり、$r$の値の大小で包装状態を判別することができる。
    }
    {
      In the above equation, $A$ and $B$ represent the point clouds of $T_{pc-pre}$ and $T_{pc-post}$ after ICP matching, respectively. Also, $a$ and $b$ represent the coordinates of each point in $A$ and $B$, and $\vec{n}_{Ai}$ represents the normal vector at point $a_{i}$ in $A$. First, for each point $b_{i}$ in $B$, $k$ points from the point cloud $A$ that are closest in distance to $b_{i}$ are selected as ($a_{i1}$, $a_{i2}$, ..., $a_{ik}$), and this set is defined as $Neighborhood(B_{i})$. Next, the normal vectors at each point in $Neighborhood(B_{i})$ are averaged to obtain $\overline{\vec{n}_{Ai}}$. Then, for each point in $B$, the angle between $\vec{n}_{Bi}$ and $\overline{\vec{n}_{Ai}}$ is calculated. The number of points for which this angle exceeds 10 degrees is counted and divided by the size of the set $B$, denoted as $M$. This yields the percentage of points where the state of the normal vectors differs significantly when comparing $T_{pc-pre}$ and $T_{pc-post}$. In other words, this indicates how well the Wrapper is wrapped around the surface of the Target at the end of the wrapping operation. The top row of \figref{evaluation-compare} shows a good example of wrapping that maintains tension in the paper along the surface of the Target. The lower row shows a bad example where the Wrapper became inflated. The values of $r$ are 0.245 and 0.344, respectively, and the wrapping state can be judged by the magnitude of $r$.
    }

\section{Experiments}
\myswitchlang
{前章までで説明した各手順について、統合実験としてそれらをつなげて一連の動作として行う実験を行う。}
{Link the procedures described in previous chapters together as a series of operations and conduct an integration experiment.}

\myswitchlang
{実験にはカワダロボティクスの開発したHIRONXに、\figref{hand-pic}のハンドを取り付けて行った。全体のRGB画像及び点群画像はロボット頭部に取り付けた深度カメラによって取得した。また、$T_{pc-pre}$及び$T_{pc-post}$はロボットの右ハンドに取り付けた深度カメラを回すことで取得した。$f^{ctrl}_{3}$の$f$の値は4 N、$f^{recog}_{5}$の式(1)における$D_{P_i}$, $D_{L_i}$, $D_{V_i}$の比重は2:3:1、$f^{ctrl}_{6}$の$thre$は5 cm, $f^{recog}_{7}$の$k$は10とした。}
{The experiments were conducted using the HIRONX, developed by Kawada Robotics, with the hand shown in \figref{hand-pic} attached. A depth camera attached to the robot's head acquired the RGB image and point cloud image. The $T_{pc-pre}$ and $T_{pc-post}$ images were acquired by rotating the depth camera attached to the robot's right hand. The value of $f$ in $f^{ctrl}_{3}$ is \SI{4}{N}, the specific weights of $D_{P_i}$, $D_{L_i}$ and $D_{V_i}$ in $f^{recog}_{5}$'s equation (1) are 2:3:1, $thre$ in $f^{recog}_{6}$ is \SI{5}{cm}, $k$ of $f^{eval}$ is 10.}

\subsection{Rectangular Shaped Box Wrapping}
\begin{figure}[thbp]
  \centering
  \includegraphics[width=0.9\columnwidth]{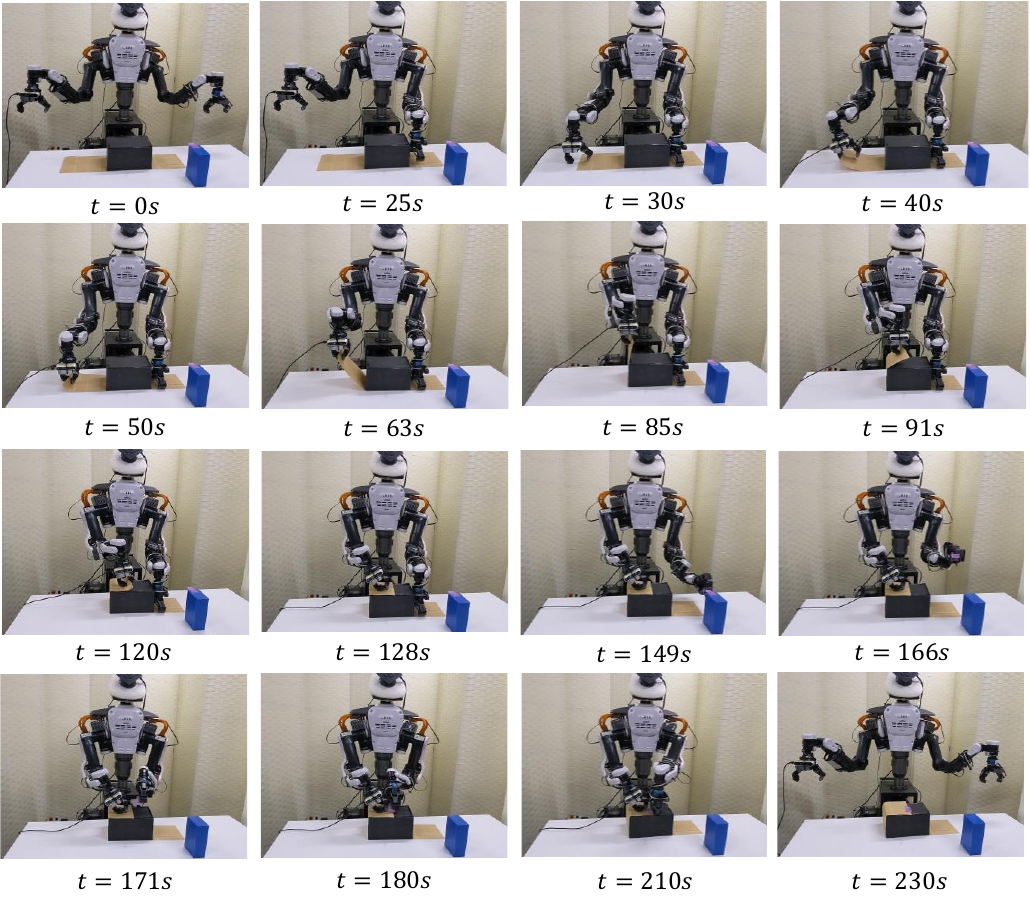}
  \caption{Experiments in wrapping rectangular boxes}
  \label{figure:experiment-box}
\end{figure}

\myswitchlang
{Targetとして縦幅約14.5 cm,横幅約21.5 cm,高さ約 11.5 cm,重量約795 gの直方体型の箱、Wrapperとして長辺60 cm, 短辺17.5 cmの包装紙、Sealとして幅3 cmのマスキングテープを用いた実験の様子を\figref{experiment-box}に示す。また認識の結果を\figref{experiment-recog}の上段に示す。$f^{recog}_{1}$においてTargetは縦14.8 cm, 横20.3 cm, 高さ10.0 cmのbboxとして認識された。また、$T_{seal-\vec{n}}$は(-0.038, -0.010, 0.999)として認識された。また、$f^{recog}_{9}$で$r$の値が0.233となった。これは\figref{evaluation-compare}上段で示す良い包装の例よりも$r$が小さい。}
{The experiment used a rectangular box of about \SI{14.5}{cm} in length, \SI{21.5}{cm} in width, \SI{11.5}{cm} in height, and \SI{795}{g} in weight as the Target, wrapping paper of \SI{60}{cm} on the long side and \SI{17.5}{cm} on the short side as the Wrapper, and masking tape of \SI{3}{cm} in width as the Seal is shown in \figref{experiment-box}. The recognition results are shown in the upper row of \figref{experiment-recog}. In $f^{recog}_{1}$, Target was recognized as a bbox with 14.8 cm height, \SI{20.3}{cm} width, and \SI{10.0}{cm} height. Also, $T_{seal-\vec{n}}$ was recognized as (-0.038, -0.010, 0.999). Also, $f^{eval}$ gave a value of $r$ of 0.233. This is smaller than the example of good wrapping shown in the upper row of \figref{evaluation-compare}.}

\subsection{Cylindrical Containers Wrapping}
\begin{figure}[thbp]
  \centering
  \includegraphics[width=0.9\columnwidth]{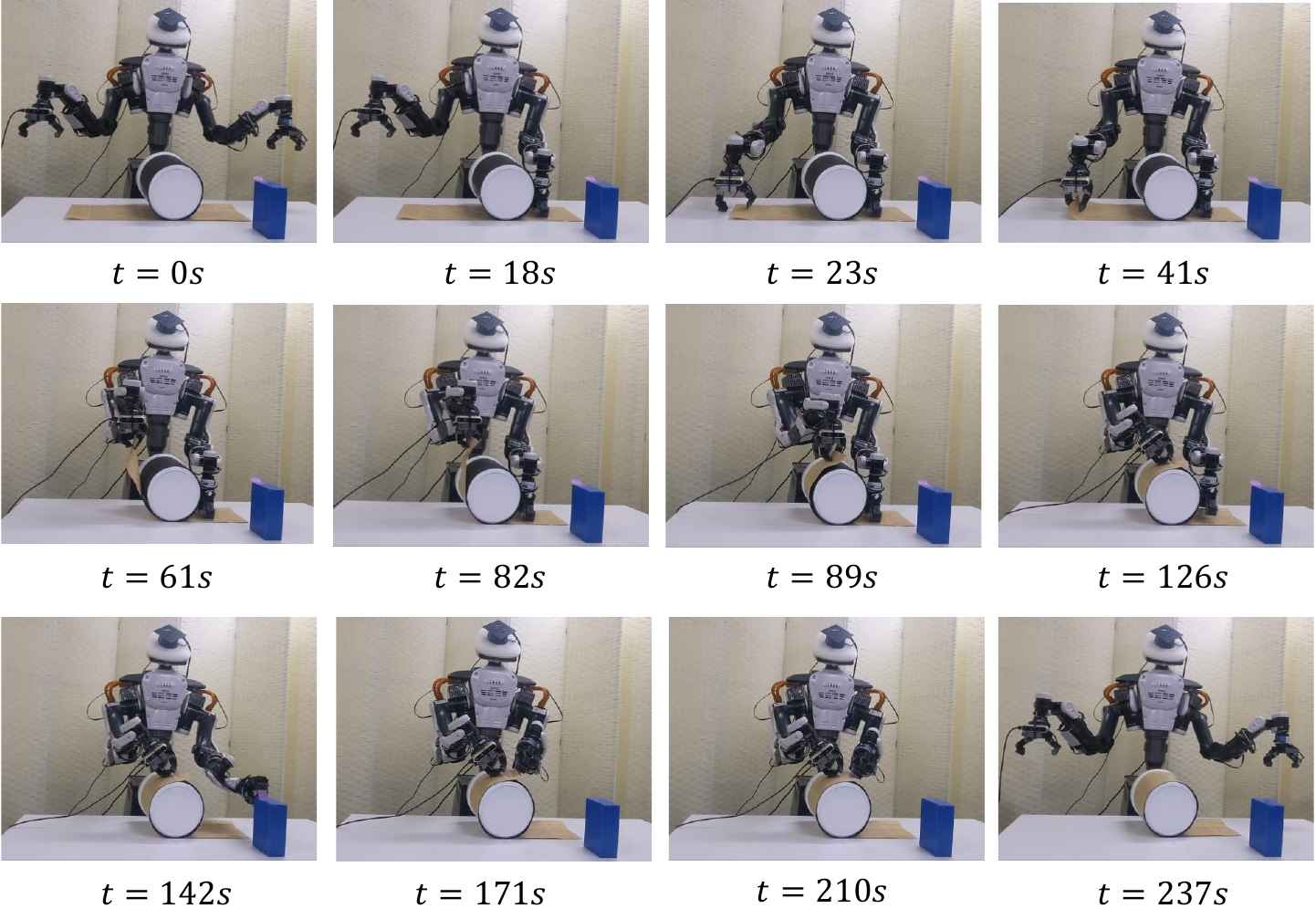}
  \caption{Experiments in wrapping cylindrical containers}
  \label{figure:experiment-cylinder}
\end{figure}

\begin{figure}[thbp]
  \centering
  \includegraphics[width=0.8\columnwidth]{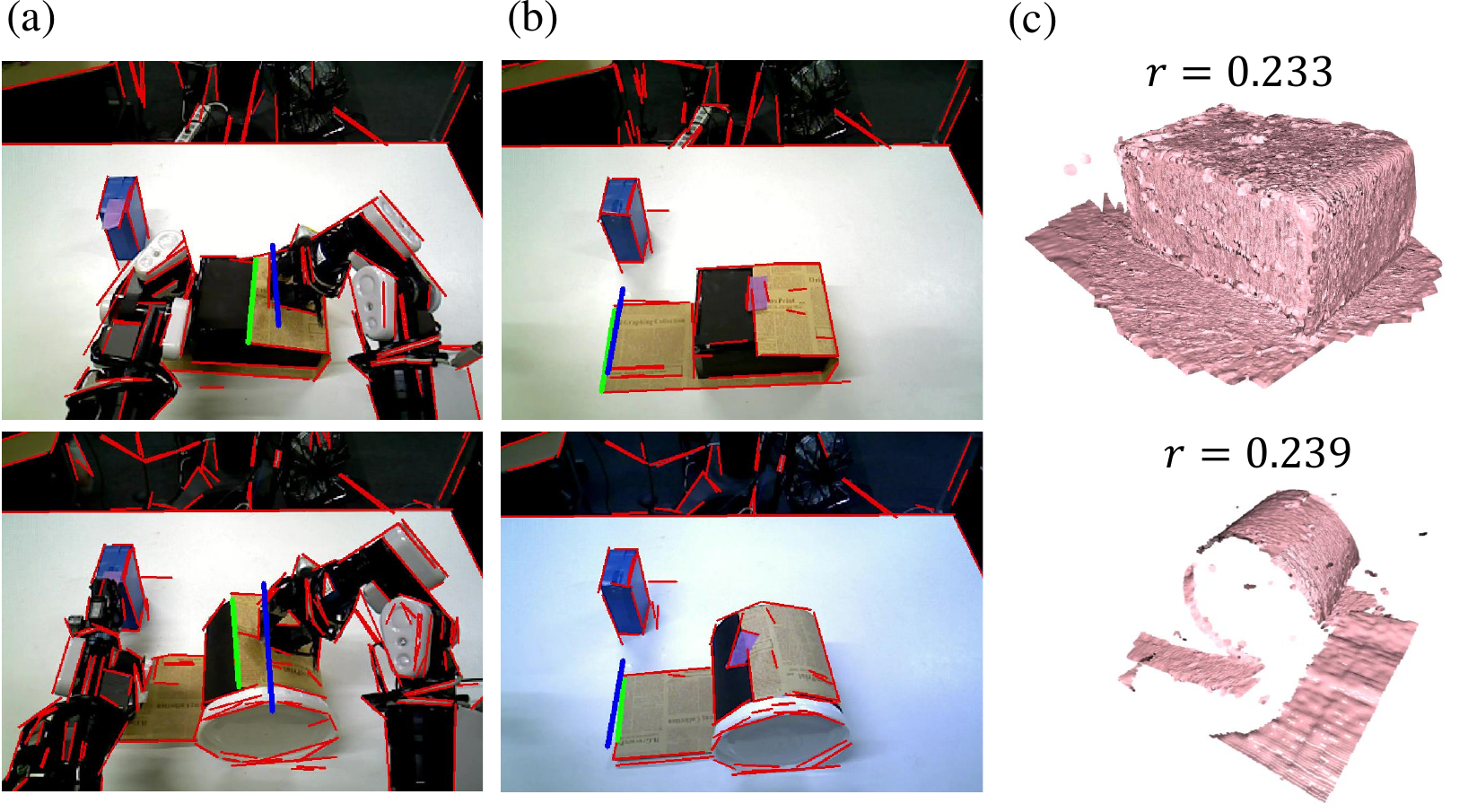}
  \caption{The upper row represents the recognition results of Experiment A. The lower row represents the recognition results of Experiment B. (a), (b), and (c) represent the recognition results for $T_{seal-pos}$, $W_{cd}$ for $f^{recog}_{5}$ of Step F, and $T_{pc-post}$, respectively. The recognition using edge similarity by $f^{recog}_{5}$ was performed correctly, and the evaluation of the value of $r$ resulted in proper wrapping.}
  \label{figure:experiment-recog}
\end{figure}

\myswitchlang
{Targetとして半径約9.0 cm, 高さ約20 cm、重量約943 gの円柱状の容器を用いた実験の様子を\figref{experiment-cylinder}に示す。また認識の結果を\figref{experiment-recog-cylinder}の下段に示す。Targetは$f_^{recog}_{1}$で縦18.6 cm, 横14.9 cm, 高さ16.0 cmのbboxとして認識された。また、$T_{seal-\vec{n}}$は(0.037, 0.344, 0.938)として認識された。また、$f^{recog}_{9}$で$r$の値が0.239となった。}
{The experiment used a cylindrical container with a radius of about \SI{9.0}{cm}, a height of about \SI{20}{cm}, and a weight of about \SI{943}{g} as the Target is shown in \figref{experiment-cylinder}. The recognition results are shown in the lower row of \figref{experiment-recog}. In $f^{recog}_{1}$, Target was recognized as a bbox of \SI{18.6}{cm} in length, \SI{14.9}{cm} in width, and \SI{16.0}{cm} in height. Also, $T_{seal-\vec{n}}$ was recognized as (0.037, 0.344, 0.938). Finally, $f^{eval}$ gave a value of $r$ of 0.239. This is smaller than the example of good wrapping shown in the upper row of \figref{evaluation-compare}.}


\section{Conclusions}
\myswitchlang
{本研究では、今まで特殊なハードウェアによる開発やある一動作の動作の生成や改善が主に行われてきたロボットによる柔軟物体操作のうち、包装動作をに取り組んだ。様々な動作行動が可能な双腕ヒューマノイドロボットによる実現や、連続的な動作の実現に注目した。包装で扱うそれぞれの物体について、その材料的特徴や包装手順過程におけるロボットによる作用・変形をもとに必要な情報を整理し記号化した。それらに基づいて包装の各手順において必要なハードウェア設計要件をまとめ、認識やマニピュレーションを入力と出力を明らかにして関数化し、システムを一般化した。また、今まで行われてきた折り紙や布の折りたたみなどと比べて立体性が増すことに着目し、アドミッタンス制御、柔軟物体の状態を保つための状態の固定や動作のつながりを考慮した両手によるマニピュレーション、点群による包装状態の評価をシステムに取り入れた。これらの統合システムを用いた統合実験では異なる形状の物品の包装行動を一連で行うことができ、システムの一般性や有効性が示された。}
{
  In this study, we addressed the wrapping operation, one of the flexible object manipulations, which has been mainly developed using specialized hardware or focused on the generation and improvement of a single action. We focused on realizing this operation using a dual-armed humanoid robot capable of performing various actions, with a particular focus on achieving continuous operations.
  For each object handled in wrapping, the necessary information was organized and coded based on its material characteristics and the actions and deformations of the robot during the wrapping procedure. Based on this information, we summarized the hardware design requirements for each procedure, clarified the inputs and outputs for recognition and manipulation, and generalized the system into a function. In addition, focusing on the increased three-dimensionality of the system compared to the conventional folding of origami and cloth, the system incorporates admittance control, two-handed manipulation that considers state fixation and motion connections to maintain the state of flexible objects and evaluation of the state of wrapping by point clouds. The experiments using these integrative systems showed that the system could perform a series of wrapping actions for objects of different shapes, demonstrating the generality and effectiveness of the system.}

\myswitchlang
{現段階では包装される物品を直方体近似して情報を保持し動作を行っているが、このシステムを基盤として、直方体近似では動作できない物体の条件や追加で保持するべき情報について調べることが今後の研究課題となりうる。また、本実験では柔軟物体の点群を色によって抽出したが、色が均一ではない柔軟物体を考慮し、新たな柔軟物の表現方法として、柔軟物体全体をグリッドで分けていくつかの点の位置姿勢情報を保持し、トラッキングなどを用いてそれらをリアルタイムで認識することが考えられる。}
{At the present stage, the information is retained and operated using a rectangular approximation of the item to be wrapped. However, using this system as a foundation, future research may be required to investigate the conditions of objects that cannot be operated using a rectangular approximation and the additional information that should be retained. In this experiment, the point cloud of the flexible object was extracted by color. However, considering flexible objects that are not uniformly colored, a new way to represent flexible objects could be to divide the entire flexible object into a grid, retain the position and posture information of several points, and recognize them in real time using tracking or other methods.}

\myswitchlang
    {
      また、本実験では紙をめくり物体の面を覆いテープを貼るという動作に取り組んだが、紙の折り目をつける、飾りをつけるなど、包装で行われる他の異なる動作に必要なハードウェア設計要件や対象物体の表現方法についても考察することも、より包括的な柔軟物体操作を行えるロボットシステムの構築に必要である。
    }
    {
      In this experiment, we focused on actions such as turning over paper to cover the object's surface and applying tape. However, to build a more comprehensive robotic system capable of flexible object manipulation, it is also necessary to consider the hardware design requirements and the representation methods for target objects needed for other wrapping-related tasks, such as creating paper folds or attaching decorations.
}

\addtolength{\textheight}{-12cm}   


\bibliographystyle{junsrt}
\bibliography{main}

\end{document}